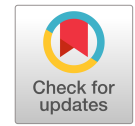

# Probabilistic Classification of Near-Surface Shallow-Water Sediments Using a Portable Free-Fall Penetrometer

Md Rejwanur Rahman, M.ASCE[1]; Adrian Rodriguez-Marek, Ph.D., M.ASCE[2]; Nina Stark, Ph.D., M.ASCE[3]; Grace Massey, Ph.D.[4]; Carl Friedrichs, Ph.D.[5]; and Kelly M. Dorgan, Ph.D.[6]

**Abstract:** The geotechnical evaluation of seabed sediments is important for engineering projects and naval applications, offering valuable insights into sediment properties, behavior, and strength. Obtaining high-quality seabed samples can be a challenging task, making in situ testing an essential part of site characterization. Free-fall penetrometers (FFPs) are robust tools for rapidly profiling seabed surface sediments, even in energetic nearshore or estuarine conditions and shallow as well as deep depths. Although methods for interpretation of traditional offshore cone penetration testing (CPT) data are well-established, their adaptation to FFP data is still an area of research. This study introduces an innovative approach that utilizes machine learning algorithms to create a sediment behavior classification system based on portable free-fall penetrometer (PFFP) data. The proposed model leverages PFFP measurements obtained from multiple locations, such as Sequim Bay (Washington), the Potomac River, and the York River (Virginia). The results show 91.1% accuracy in the class prediction, with the classes representing cohesionless sediment with little to no plasticity (Class 1), cohesionless sediment with some plasticity (Class 2), cohesive sediment with low plasticity (Class 3), and cohesive sediment with high plasticity (Class 4). The model prediction not only predicts classes but also yields an estimate of inherent uncertainty associated with the prediction, which can provide valuable insight into different sediment behaviors. Lower uncertainties are more common, but they can increase significantly depending on variations in sediment composition, environmental conditions, and operational techniques. By quantifying uncertainty, the model offers a more comprehensive and informed approach to sediment classification. **DOI: 10.1061/JGGEFK.GTENG-13486.** *This work is made available under the terms of the Creative Commons Attribution 4.0 International license, https://creativecommons.org/licenses/by/4.0/.*

## Introduction

The use of free-fall penetrometers (FFPs) is motivated by the need for rapid and quantitative geotechnical assessment of seabed sediments in a variety of engineering projects and naval applications. In situ tests to obtain geotechnical properties are an important offshore site investigation method due to the inherent challenges associated with collecting high-quality seabed samples in many environments. Free-fall penetrometers can swiftly profile the seabed's surface sediments, even in dynamic nearshore conditions and to depths ranging from several meters to over 1,000 m below the water's surface (Akal and Stoll 1995; Albatal and Stark 2017). FFP instruments collect data with a vertical resolution often better than 1 cm in the uppermost seabed layer. Moreover, they can be deployed in a cost- and time-efficient manner in settings characterized by energetic hydrodynamics and active sediment dynamics (Stark et al. 2012; Albatal et al. 2020). Data sets obtained from FFP tests have found applications in various domains. For example, Mulukutla et al. (2011) used FFP data to infer information about the grain size of the sediments. Aubeny and Shi (2006), Chow et al. (2014, 2017) estimated undrained shear strength based on FFP measurements in fine-grained sediments, and Albatal et al. (2020) established correlations between FFP decelerations and relative density and explored diverse methods for estimating critical friction angles in coarse-grained sediments. Jaber and Stark (2023) introduced a framework to obtain geotechnical properties such as undrained shear strength or friction angle for any typical seabed sediment types.

A wide range of methodologies has been proposed for classifying data derived from cone penetration testing (CPT). These methodologies encompass both deterministic approaches (e.g., Robertson 1990; Olsen and Mitchell 1995), which provide fixed classifications, and probabilistic techniques (e.g., Jung et al. 2008; Cetin and Ozan 2009), which account for uncertainties in soil behavior. Additionally, Pradhan (1998) developed fuzzy membership functions based on the Robertson (1990) chart, and Zhang and Tumay (1999) introduced a fuzzy soil classification method derived from the Douglas and Olsen (1981) chart to account for the uncertainty and imprecision. Although these advancements have significantly improved traditional CPT classification techniques, the application of probabilistic approaches to free-fall penetrometer data remains largely unexplored. In addition, balancing simplicity in the classification process with the need to account for inherent uncertainties also is essential for developing robust and reliable methodologies. Probabilistic machine learning techniques, in particular, offer significant potential for addressing the variability and uncertainty associated with FFP data, paving the way for more a more flexible sediment classification framework. Hence, this study had the following goals:
1. To develop a sediment behavior classification scheme using portable FFP (PFFP) data; and
2. To quantify associated uncertainties in sediment behavior classification.

[1]Ph.D. Candidate, Charles E. Via, Jr. Dept. of Civil and Environmental Engineering, Virginia Tech, 750 Drillfield Dr., Blacksburg, VA 24060 (corresponding author). ORCID: https://orcid.org/0000-0002-1954-3561. Email: rejwan@vt.edu

[2]Professor, Charles E. Via, Jr. Dept. of Civil and Environmental Engineering, Virginia Tech, 750 Drillfield Dr., Blacksburg, VA 24060. ORCID: https://orcid.org/0000-0002-8384-4721. Email: adrianm@vt.edu

[3]Associate Professor, Engineering School of Sustainable Infrastructure and Environment, Univ. of Florida, Gainesville, FL 32611. Email: nina.stark@essie.ufl.edu

[4]Research Scientist, Virginia Institute of Marine Science, William & Mary, Gloucester Point, VA 23062. ORCID: https://orcid.org/0000-0001-7936-1586. Email: grace.massey@vims.edu

[5]Professor, Virginia Institute of Marine Science, William & Mary, Gloucester Point, VA 23062. ORCID: https://orcid.org/0000-0002-1810-900X. Email: carl.friedrichs@vims.edu

[6]Associate Professor, Univ. of Texas at Austin–Marine Science Institute, Port Aransas, TX 78373. Email: kelly.dorgan@austin.utexas.edu

This study proposes a predictive model based on machine learning. The model leverages portable free-fall penetrometer (PFFP) measurements obtained from Sequim Bay (Washington) and the Potomac and York Rivers (Virginia). This paper begins by discussing the data collection procedure and corresponding lab testing. Then different technical components (i.e., model architecture, data set processing, and model training) of the proposed model are presented. Next, a step-by-step guide for predicting the sediment class using the proposed model is presented. Then the results of the model are presented and discussed, along with some of the example prediction scenarios. In addition, a qualitative discussion is presented of how the results obtained from this model can be used to predict sediment behaviors.

## Data Collection

### Deceleration, Velocity, and Penetration Depth

The PFFP used in this study was equipped with a set of five accelerometers designed to measure both vertical deceleration and tilt, allowing the derivation of vertical velocity and bed penetration depth. These accelerometers had varying measurement ranges, from ±2 to ±250 $g$, where $g$ is gravitational acceleration. The PFFP was 63.1 cm in length and 8.75 cm in diameter, and weighed 7.7 kg when equipped with the 60° conical tip option. Several studies have investigated the processing of accelerometer-based PFFP data (e.g., Mulukutla et al. 2011; Albatal and Stark 2017; Mumtaz et al. 2018; Bilici et al. 2019; Hunstein et al. 2023). By integrating deceleration data recorded during seabed penetration, the probe's velocity and depth of penetration were calculated. Fig. 1(a) shows a PFFP deployment, and Figs. 1(b and c) present PFFP sample data displaying the measured deceleration and derived penetration velocity versus its displacement.

### Survey Locations and Sediment Sampling

Surveys were conducted in three distinct locations—Sequim Bay, WA; Potomac River, VA; and York River, VA—between 2018 and 2022. The locations are shown in Fig. 2, and a detailed distribution of the number of PFFP deployments at different locations is presented in Table 1. Within the York River estuary, deployments were divided into eight sublocations. In total, 447 PFFP deployments were carried out at these three sites. In addition to PFFP testing, sediment samples and cores were collected using various techniques based on the specific sediment types and with the goal of retrieving high-quality seabed samples. For cohesive sediments, methods such as box coring and gravity coring were employed, whereas diver sampling primarily was used in sandy sediment areas, in addition to grab sampling.

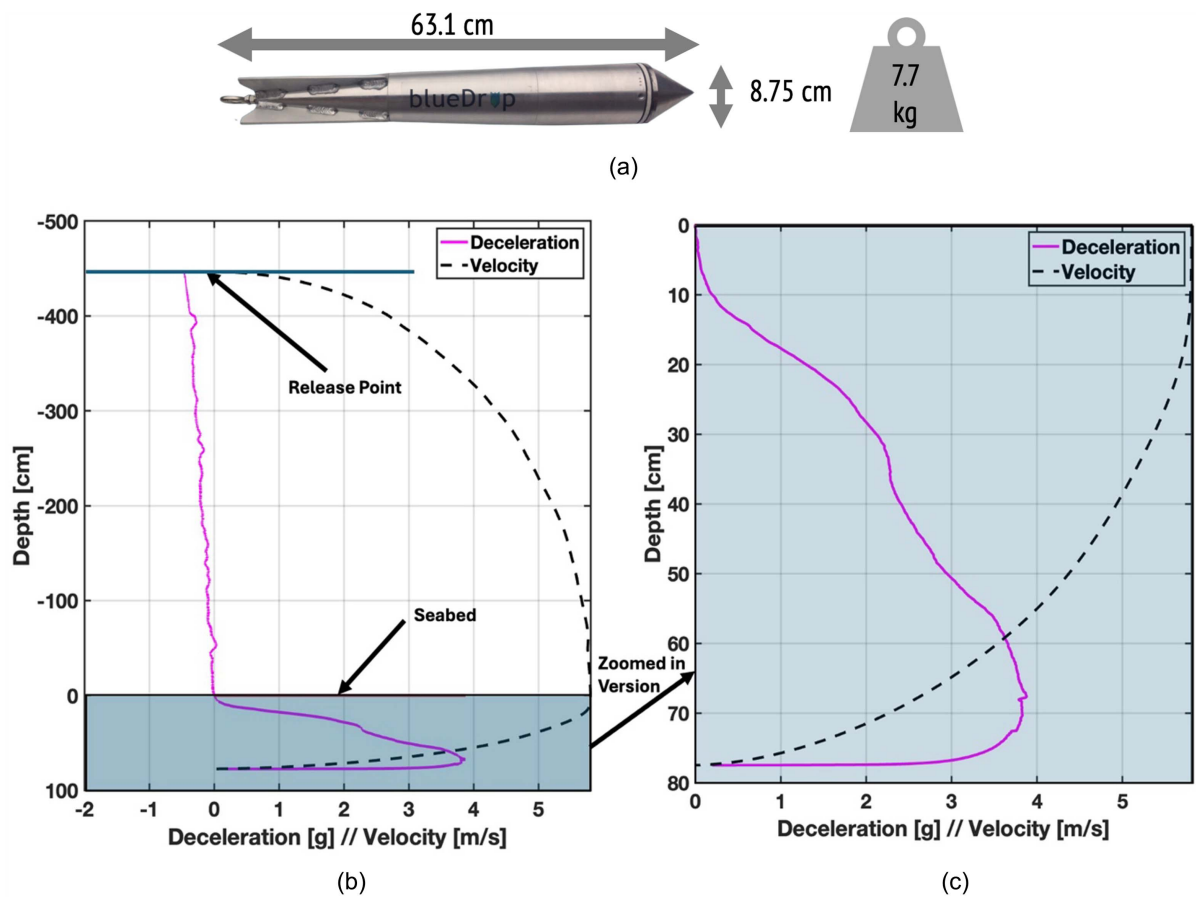

**Fig. 1.** (a) Dimensions and weight of PFFP (image courtesy of Blue C Designs Ltd.); (b) displacement versus deceleration and velocity plots of PFFP from point of release to stop; and (c) magnified version of Fig. 1(b) showing seabed penetration only.

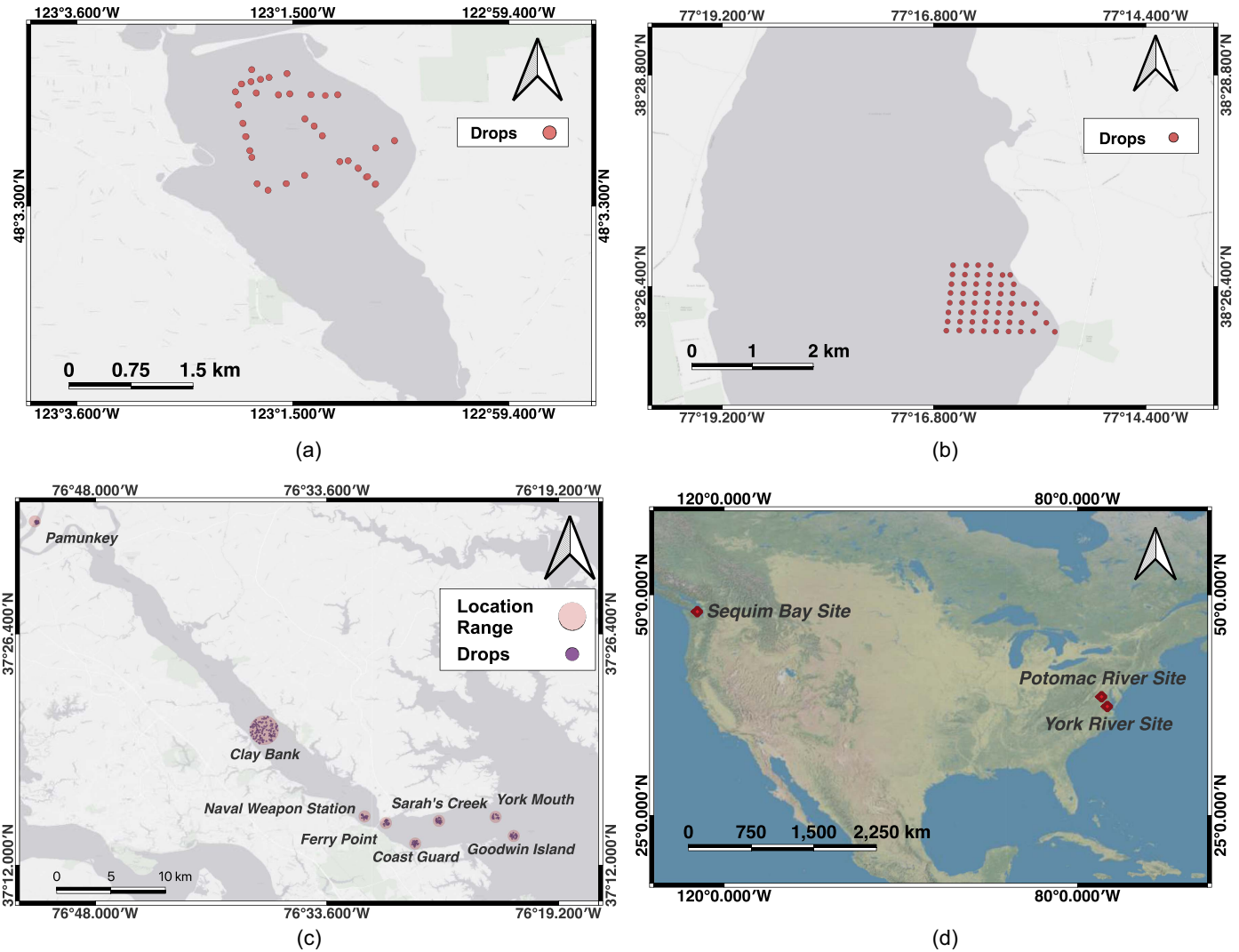

**Fig. 2.** PFFP deployment locations: (a) Sequim Bay; (b) Potomac River; (c) York River; and (d) overview map. [Base maps (a–c) from World Light Gray Base Map, Sources: Esri, HERE, Garmin, © OpenStreetMap contributors, and the GIS User Community; base map (d) from ArcGIS World Physical Map.]

## Sediment Class Determination

The collected sediment samples were classified according to the Unified Soil Classification System (USCS) outlined in ASTM D2487 (ASTM 2017). This classification system considers particle-size gradation and Atterberg limits, including the plasticity index and liquid limit. Based on the USCS classification, the sediments were categorized into four major types. Detailed information regarding these soil behavior types is presented in Table 2. The sand content (percentage of sediment with grain size larger than 0.075 mm) and fines content (percentage of sediment with grain size smaller than 0.075 mm) were determined in accordance with the specifications outlined by the USCS (ASTM D2487). Classes 1 and 2 can be categorized as sandy sediments with a sand content exceeding 50%. The primary distinction between Class 1 and Class 2 is based on the fines content, which is set at 12% according to the Unified Soil Classification System. Similarly, Classes 3 and 4 are classified as fine-grained muddy sediments, characterized by a fines content greater than 50%. The key differentiating criterion between Class 3 and Class 4 is the liquid limit. It has been found that clay soils composed of highly plastic clay minerals such as montmorillonite tend to have higher liquid limits than low-plasticity clay minerals such as kaolinite (Mitchell and Soga 2005). Therefore, the liquid limit can be used as a proxy for determining plasticity of a sediment. The USCS distinguishes high and low plasticity of clay or silt by a threshold value of liquid limit equal to 50. The same criteria are used herein to distinguish between Class 3 and Class 4. The sediment

**Table 1.** Distribution of PFFP data set among different locations

| Location | Sublocation | Number of PFFP deployments |
|---|---|---|
| York River, VA | Clay Bank | 126 |
| | Pamunkey | 45 |
| | Naval Weapon Station | 25 |
| | Ferry Point | 21 |
| | Sarah's Creek | 28 |
| | Coast Guard | 25 |
| | York Mouth | 15 |
| | Goodwin Island | 27 |
| Potomac River, VA | — | 72 |
| Sequim Bay, WA | — | 63 |
| Total | — | 447 |

Table 2. Gradation and plasticity criteria for sediment behavior types used in this study

| Class No. | Sediment behavior types | Corresponding USCS class | Data set size |
|---|---|---|---|
| 1 | Cohesionless sediment with little to no plasticity: sand content > 50% and fines content < 12% | SP, SW, SP-SM, SP-SC, SW-SM, SW-SC | 92 |
| 2 | Cohesionless sediment with some plasticity: sand content > 50% and fines content > 12% | SM, SC, SC-SM | 65 |
| 3 | Cohesive sediment with low plasticity: fines content > 50% and liquid limit < 50% | ML, CL, CL-ML | 23 |
| 4 | Cohesive sediment with high plasticity: fines content > 50% and liquid limit > 50% | MH, CH | 267 |

Note: SP = poorly graded sand; SW = well-graded sand; SM = silty sand; SC = clayey sand; ML = low-plastic silt; CL = low-plastic clay; MH = high-plastic silt; and CH = high-plastic clay.

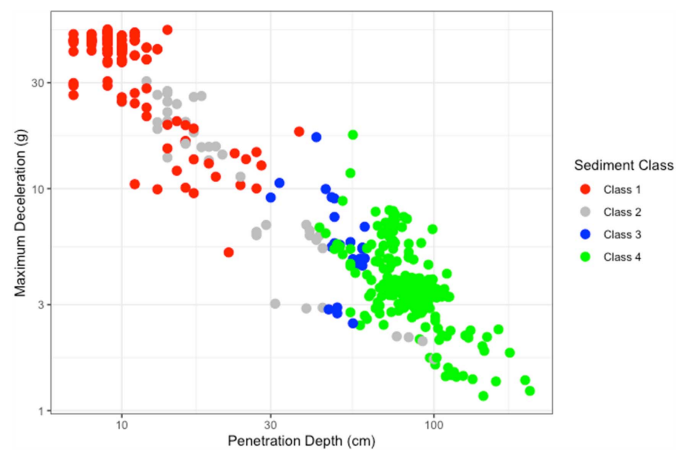

**Fig. 3.** PFFP data used for training and testing of the predictive model displayed in terms of the maximum deceleration recorded by the PFFP, the PFFP penetration depth, and the assigned sediment classes based on laboratory testing of sediment samples.

class obtained from laboratory testing then was used to train the predictive model for connecting the PFFP data to the sediment classes. Among the four classes, Class 4 was identified as the most dominant, comprising a total of 267 data points. This was followed by Class 1, with 119 data points; Class 2, with 40 data points; and Class 3, with 23 data points. The issue of data set imbalance is addressed subsequently. The data set employed in this study are mapped onto the space of maximum deceleration and penetration depth in Fig. 3.

## Probabilistic Predictive Model

We used a probabilistic predictive model approach, which builds on previous work which utilized subsets of parameters obtained from the PFFP data. Jaber and Stark (2023) classified sediments into cohesionless versus cohesive solely based on PFFP penetration depth. Mulukutla et al. (2011) suggested a normalized parameter named firmness factor (which is derived from maximum deceleration, impact velocity, and time of penetration) for classifying the sediments into different grain size classes. These classification schemes focus on specific values of the deceleration curves (such as maximum deceleration) rather than using the full deceleration curves. This leaves information unused, and may lead to a higher chance of misclassification. Our study considers the full deceleration curve during seabed penetration, and uses the previous prediction approaches as prior knowledge. Hence, the prediction scheme is divided into two steps:

1. prediction of sediment class using normalized maximum deceleration and penetration depth; and
2. prediction of sediment class using a full and normalized deceleration curve within a Bayesian framework and using the results of Step 1 as a prior, in which normalized deceleration is defined as

$$\text{Normalized deceleration} = \frac{\text{Deceleration}}{\text{Impact velocity}} \quad (1)$$

Machine learning was chosen as the modeling approach in this study. This choice was motivated by several factors. Over the last decade, machine learning has emerged as a leading methodology in predictive modeling. In particular, we adopted a probabilistic machine learning approach, which aligns with the frameworks commonly employed in other scientific and engineering disciplines (Murphy 2022). This approach offers better robustness against overfitting on small data sets than traditional approaches (Gal and Ghahramani 2016). Furthermore, after the model architecture is established, it can be extended seamlessly to incorporate additional training data, enabling continuous improvement.

In the first step, a random forest model was employed. Random forest, which is a tree-based machine learning algorithm, is used widely due to its robustness and ability to perform well without extensive hyperparameter tuning (Müller and Guido 2016). This model used normalized maximum deceleration and penetration depth as inputs for sediment classification. For the second step, a one-dimensional (1D) Bayesian convolutional neural network (CNN) was applied, using the full deceleration curve as the input. The full deceleration curve contains a large number of correlated deceleration values, and CNNs are particularly well-suited for handling such data while preserving their interdependencies. The 1D CNN model was chosen to extract features from the entire deceleration curve and thus improve classification accuracy. Both of these models have inherent limitations. The random forest model primarily focuses on penetration depth and maximum deceleration, neglecting other informative features of the deceleration curve. Conversely, the 1D CNN model excels in identifying features across the deceleration curve's shape but does not account for overall penetration depth, which is a critical parameter for sediment classification. To overcome these limitations, these two methods were combined into a robust prediction framework that integrates complementary strengths, offering a comprehensive approach to sediment classification. A schematic diagram of the proposed predictive model is shown in Fig. 4.

### Random Forest Model

Random forest (Breiman 2001), a versatile and powerful ensemble decision-tree technique commonly utilized in machine learning, was used to make our initial prediction. It combines the predictive power of multiple decision trees to improve the overall accuracy and robustness of predictions. A decision tree aims to select the best feature at each node to maximize information gain or minimize impurity. In other words, it identifies features that split the data into groups that are as different from each other as possible, or that make the groups as pure and similar as possible. In a random forest, a collection of decision trees is constructed, each trained on a random subset of the data set and with the ability to select a random subset of features for each split. This randomness mitigates overfitting and promotes generalization. In the prediction phase, each tree in the ensemble yields a prediction. These predictions then are combined, either by voting or averaging, to form a single final prediction. This process makes the model more accurate than a single decision-tree prediction, and less affected by any irregularities or extreme values in the data.

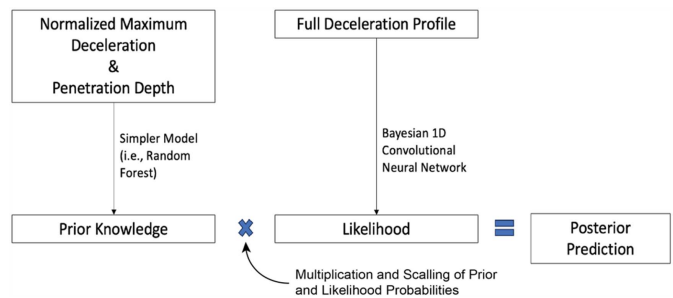

**Fig. 4.** Schematic diagram of the proposed predictive model.

In the present case, the normalized maximum deceleration and penetration depth were fed into a random forest model to obtain a probability estimate of the corresponding sediment classes. The model was implemented using the Scikit-Learn version 1.5.1 package (Pedregosa et al. 2011) in Python. At first, 15% of the full data set was kept aside for use as a test data set. The rest of the data were split into 85% training and 15% validation subsets. A grid search with fivefold cross-validation was employed to identify the optimal hyperparameters, including the number of estimators, maximum tree depth, minimum samples per split, and leaf nodes. The effect of bootstrap sampling (with and without replacement) also was included in the grid search to evaluate its impact on model performance. The best model, selected based on validation accuracy, was trained further on the full training data set.

### Bayesian 1D Convolutional Neural Network

Neural networks can take in a large number of input variables, and have been used extensively in the past in applications related to classification (Krogh 2008). In particular, CNNs have been found to be successful and efficient in the analysis and classification of structured grid data such as images, audio, time series, and text sequences, and commonly are used in tasks such as image classification, object detection, and segmentation. For sequential data, such as time series or text, the use of a 1D CNN generally is recommended (Kiranyaz et al. 2021). We adopted a Bayesian variant of a CNN, which offers better robustness against overfitting on small data sets than traditional approaches (Gal and Ghahramani 2016). The model used for this study uses the full seabed penetration deceleration curve and provides the output as uncertainty estimates of each sediment class. A diagram showing the components of the 1D CNN used in this study is shown in Fig. 5.

In the initial layer of a CNN, the network traditionally is tasked with handling the input sequence. In our specific context, we gathered deceleration data at a high frequency of 2 kHz time series (i.e., captured at 0.005-s intervals). However, the size of this input exceeded the number of our sample data, which may have resulted in overfitting of the model. To address this, we computed the average deceleration values for every 1 cm of penetration, effectively reducing the input size while preserving the fundamental shape of the deceleration curve. As with the random forest model, this training data set also was partitioned into 85% for model training and 15% for validation to assess performance during the training process.

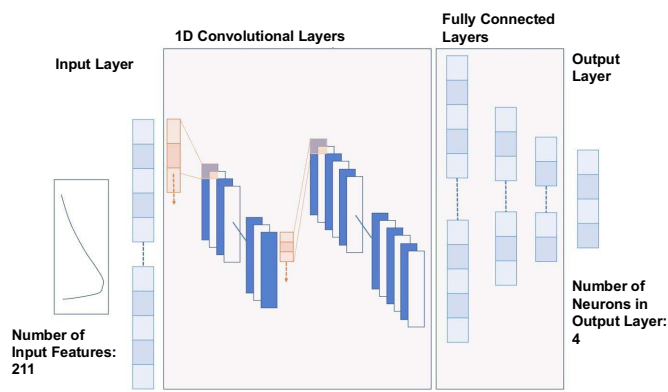

**Fig. 5.** Neural network architecture with 211-feature input layer, 1D convolutional layers for feature extraction, fully connected layers for flattening the extracted features, and 4-neuron output layer for sediment classification.

The convolutional layers within a CNN play a pivotal role in leveraging learnable filters for local convolutions. Our model utilizes a 1D CNN with two convolutional layers followed by fully connected layers for multiclass classification. To introduce nonlinearity and enhance learning capacity, we use the rectified linear unit (ReLU) activation function (Nair and Hinton 2010). After the convolutional layers, the output is flattened into a vector and passed through two fully connected hidden layers. The final output layer, with four neurons (matching the number of classes), employs the SoftMax activation function (Bridle 1989) to produce class probabilities.

In the domain of deep learning, Bayesian neural networks offer a distinct probabilistic approach. In contrast, standard neural networks operate deterministically, utilizing fixed model parameters derived from training data to provide point estimates in predictions. Although effective, these standard networks are susceptible to overfitting and lack the capability to quantify prediction uncertainty (Goodfellow et al. 2016). Instead, Bayesian neural networks treat model parameters as probability distributions. This unique approach equips them with the capacity to provide uncertainty estimates in predictions. Variational inference techniques, commonly applied in Bayesian neural networks, approximate these probability distributions, enhancing the robustness of the model and enabling more effective regularization (Graves 2011; Blundell et al. 2015). Bayesian neural networks are particularly advantageous when dealing with limited data and applications that demand a comprehensive understanding of model uncertainty. When predicting, Bayesian neural networks sample the model parameters from the trained probability distribution. This is why the model provides a different outcome each time. By creating an ensemble of these different outcomes, the uncertainty of the predictions can be estimated.

### Data Set Preprocessing

Prior to inputting the data into the model, it was necessary to perform preprocessing to ensure consistency and suitability for accurate model training. We gathered a total of 447 samples (pairs of colocated PFFP and sediment sample data). This data set is presented in Rahman (2025), and is available at the GitHub link provided in the "Data Availability" section of the present paper. From this data set, 15% of the samples were deliberately set aside randomly to form the testing data set for subsequent evaluation. Efforts were made to ensure that the testing data set included samples representing all four classes. Prior to both training and testing, all data underwent a scaling process to enhance uniformity and consistency in feature representation. However, although these steps were intended to improve data quality, perfect comprehensiveness and uniformity cannot be guaranteed. Despite the overall sample size of 447, our data set exhibited an inherent class imbalance (Table 1). Various strategies exist for managing imbalanced data sets, and we employed the adaptive synthetic sampling (ADASYN) algorithm (He et al. 2008). ADASYN is an oversampling technique specifically tailored to address imbalanced classification problems. This approach, developed for time-series data, is appropriate for our data set in which deceleration values varied across depth instead of time. This algorithm addresses the relative dependency of deceleration between different depth while oversampling. This is why this specific algorithm was used instead of simple bootstrapping techniques. ADASYN should be applied only to the training data set to maintain the reliability of testing accuracy metrics. This algorithm was used only to perform oversampling before using the 1D Bayesian CNN model. Because the random forest model had only two input parameters, using simple oversampling algorithm was sufficient.

### Combining Random Forest and 1D Bayesian CNN Model

The final task was to combine the predictions obtained from the random forest and 1D Bayesian CNN models. To do this, a Bayesian formulation has been applied. The random forest model prediction is based on the assumption that the information on the whole deceleration profile can be represented by specific points, such as maximum deceleration, penetration depth, and impact velocity (Jaber and Stark 2023). This assumption simplifies the prediction approach, yet still provides a good approximation of the sediment class. Here, the prediction obtained from the random forest model was considered as a prior probability for the 1D Bayesian CNN model. The 1D Bayesian CNN model uses the whole deceleration profile to make a prediction. The algorithm tries to identify differentiating features of the deceleration profile and predict the sediment classes based on the obtained features. This algorithm focuses equally on every part of the deceleration profile, and can reveal additional important information that was missed by the random forest approach. The prediction obtained from this approach can be considered as the likelihood in the Bayesian formulation. By multiplying these two probabilities, we can obtain the final prediction.

Because the probability values of the prior prediction are based on a simple calculation, zero probabilities may be assigned to certain classes, potentially causing issues during the Bayesian updating step. This zero-probability in the prior will make the final probability zero even if the probability obtained from the second approach is significantly higher than zero. To address this issue, the prior probabilities of the four classes were scaled between 0 and 0.6 and then a bias term of 0.1 was added to each class probabilities. These numbers were determined through trial and error to avoid extreme prior probabilities while preserving the overall order of prior probabilities among different classes. Alternatively, this bias term can be thought of as a controlling weight that is added to the random forest model. Because four classes exist, the bias term can vary only between 0 and 0.25. The larger the bias term, the less weight there is on the prior (i.e., the random forest). A value of 0.1 provides a good balance between reducing extreme probabilities and preserving relative trends among different class probabilities obtained from random forest model. This process is illustrated in Fig. 6. In this specific example, the obtained probabilities from the random forest model for Classes 1, 2, 3, and 4 are 0.8, 0.2, 0.0, and 0.0, respectively. These probability values were converted to 0.58, 0.22, 0.1, and 0.1 after scaling and adding the bias. This conversion can mitigate extreme prior probability values. As mentioned previously, the 1D Bayesian CNN model provides a different probability estimate for the same inputs each time the model is run. Therefore, running the model several times for the same input can create a range of possible probability values for each class. Based on trial and error, it was found that running the model 30–50 times provides stable bounds on the probability of different classes. Fig. 7 provides an example of a specific prediction. It displays the predicted uncertainty bounds ($Q_1$, $Q_2$, and $Q_3$) for Class 1 in this prediction after each iteration. This demonstrates that a stable prediction bound can be achieved after 30–50 iterations.

A step-by-step procedure to obtain predictions of sediment classes is outlined in Table 3. Detailed Python code (including the read me file and data) is presented in Rahman (2025). By following these steps, nonexpert users can derive an estimate of sediment behavior classes with their associated uncertainties.

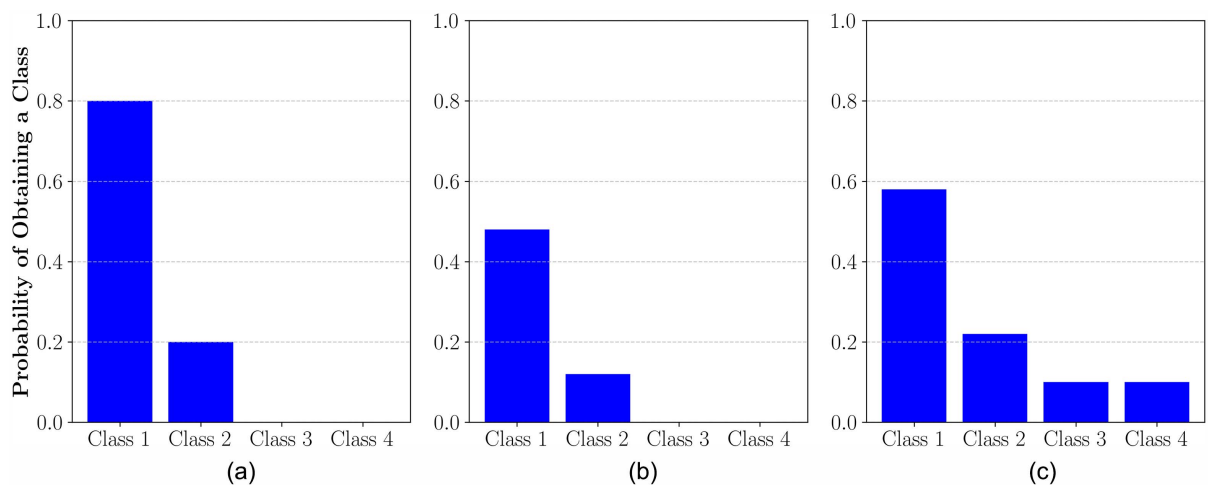

**Fig. 6.** Modification process of probability values obtained from random forest model to eliminate extreme priors: (a) probability obtained from random forest model; (b) scaled probability within 0.0–0.6 range; and (c) modified probability after adding bias term to the scaled probability.

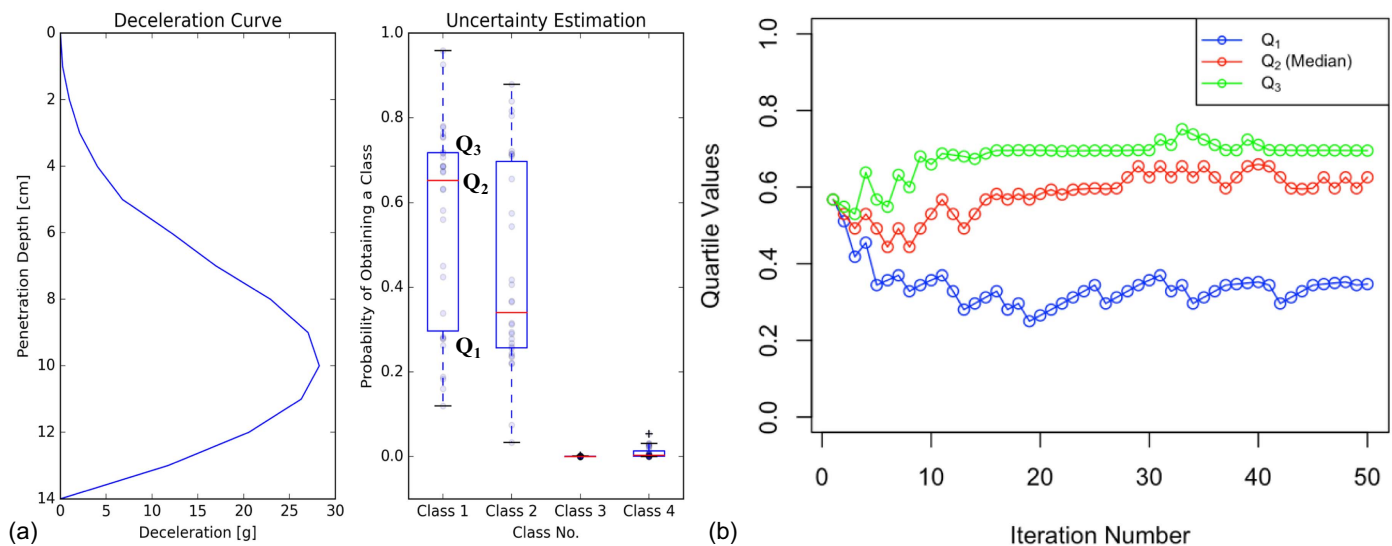

**Fig. 7.** (a) Example prediction of the model indicating the uncertainty bounds $Q_1$, $Q_2$, and $Q_3$ for Class 1; and (b) change of uncertainty bounds $Q_1$, $Q_2$ and $Q_3$ with number of iterations.

**Table 3.** Step-by-step procedure to obtain sediment behavior classes from PFFP data

| Step | Task |
|---|---|
| 1 | Obtain the deceleration profile from the start of penetrating the sediment. Integrate it with time to obtain the velocity profile and penetration depth. |
| 2 | Determine the impact velocity and obtain the normalized deceleration profile using Eq. (1). |
| 3 | Take the normalized maximum deceleration and penetration depth and feed it into the random forest model to obtain the prior probability estimate for each sediment class. |
| 4 | Take the normalized deceleration profile and extract the values at 1-cm depth intervals. Then feed the decimated profile into the 1D Bayesian CNN model to obtain the likelihood probabilities for each class. |
| 5 | Multiply each of the likelihood probabilities (Step 4) with the scaled prior probabilities (Step 3). Normalize the probability values so that the summation of probability of four classes becomes 1. |
| 6 | Repeat Steps 4 and 5 30–50 times with the same input deceleration profile. The model should provide different probability outputs for each run. Thus, it will provide a range of probabilities for each class. |
| 7 | Plot the values and create a box plot for better visualization. |

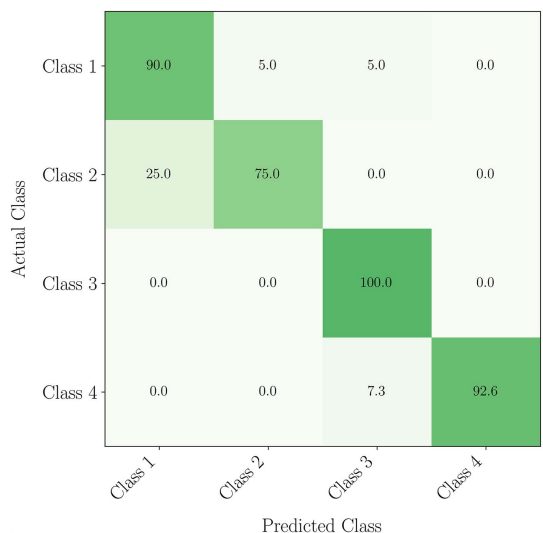

**Fig. 8.** Confusion matrix for testing data set. Percentage values on the diagonal represent the percentage of times that the predicted value matched the actual values for each class.

## Results and Discussion

The proposed classification includes uncertainty estimates. Instead of obtaining a single probability value, the model outputs the range of probability of a given class. Knowledge of the range of probabilities for different outcomes can help users assess the potential risks associated with each prediction. This allows for more informed decision-making.

The confusion matrix for the testing data set summarizes how often the classification model incorrectly assigned results from a PFFP drop to a given class (Fig. 8). The predicted class label was assigned to the class with highest median probability whereas the actual class was determined by lab test results. For example, in the case of Class 2, the model accurately predicted 75% of the samples actually belong to Class 2, and misclassified 25% as Class 1. Overall, the model was 91.1% accurate on the test data set.

Although the performance of the model can always be improved by adding additional training data sets, the authors believe that the current state of the model satisfactorily classifies the sediment and can be used in practical survey. To validate this assertion, a blind test data set from an entirely different site—Sydney Harbor, Nova Scotia, Canada—was utilized. At this location, the PFFP was deployed, and sediment samples were collected from three distinct locations (Fig. 9). The sediment classifications predicted by the model subsequently were compared with laboratory-tested results. A total of 34 PFFP deployments were analyzed, and three distinct sediment types were identified: Class 1, Class 2, and Class 3; there were no observations of Class 4 sediments. The model successfully predicted 30 of the 34 samples, achieving an accuracy of 88.2%. A detailed class-specific confusion matrix is provided in Fig. 10.

Examining the misclassifications revealed a consistent pattern: all misclassified instances of Class 1 were erroneously identified as

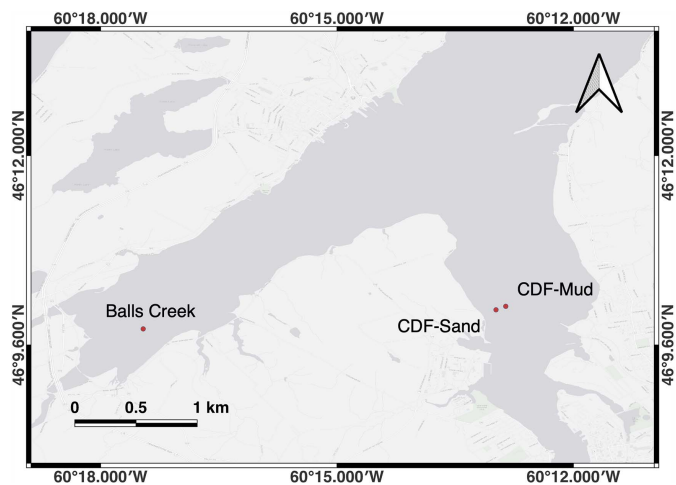

**Fig. 9.** Location of PFFP deployment at additional testing site in Sydney Harbor, Nova Scotia. (Base map from World Light Gray Base Map, Sources: Esri, HERE, Garmin, © OpenStreetMap contributors, and the GIS User Community.)

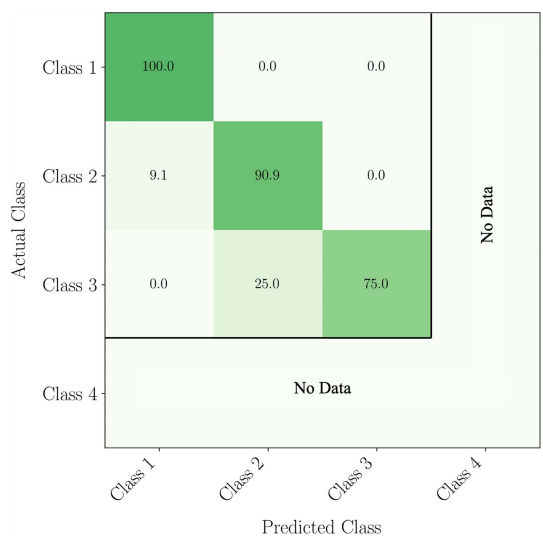

**Fig. 10.** Confusion matrix for blind testing data set. Percentage values on the diagonal represent the percentage of times the predicted value matched the actual values for each class.

Classification System. It incorporates flexible boundaries to distinguish between coarse and fine-dominant soils, accounts for the shape of coarse-grained particles, evaluates the influence of fines on mechanical and hydraulic properties, and considers the impact of pore fluid chemistry. This flexible boundary between coarse soils and fine-dominant soils approach also agrees with Whitehouse (2000), who stated that the controlling boundary is dependent on the mineralogy of sand and clay particles. Although this may help the increase the accuracy of the model, classifying the soil based on the RSCS requires determining some additional parameters such as grain roundness and liquid limit using brine and kerosene. The authors intend to address this issue in future publications.

To elaborate on the model's prediction capability, some examples are described herein. Fig. 11(a) shows an example of a confident prediction. The result shows that Class 1 (cohesionless sediment with little to no plasticity) had the highest probability of occurring with this type of deceleration curve with high certainty. This prediction matched the actual lab-tested result. Jaber and Stark (2023) suggested that a penetration depth limited to 9 cm indicates sand, also matching our predicted result. In Fig. 11(b), the uncertainty estimate has a large uncertainty between the two cohesionless classes (i.e., Classes 1 and 2), but a low chance for cohesive classes (i.e., Classes 3 and 4). This indicates that the model was very confident about the fact that it is a cohesionless sediment, but could not determine whether it belonged to Class 1 or Class 2. A different type of case is presented in Fig. 11(c). For this case, the value of the penetration depth was in an intermediate zone between the value expected for different sediment types. Thus, it is difficult to predict with certainty the class of the sediment. Consistent with this, the proposed model had large uncertainty (i.e., a spread-out box plot) in the classification. The case in Fig. 11(d) shows the benefits of using the full deceleration profile for prediction. From the traditional approach of PFFP data processing, this deployment was classified as very soft plastic sediment because the penetration depth was close to 80 cm. However, this drop clearly was different from typical PFFP drops, especially in the upper part. Because of this irregularity, the proposed model indicated some probability that it was Class 1 or 2. This also may indicate the possibility of some other issues, such as layered sediments or biological activity in the area.

The methodology presented in this paper takes a different approach from those of previous researchers (i.e., Jaber and Stark 2023; Mulukutla et al. 2011; Stark et al. 2024) for interpreting PFFP data. Instead of predicting specific geotechnical properties, such as undrained shear strength or friction angle, it classifies the deceleration curve obtained from PFFP and assigns sediment behavior types based on that classification. This method avoids imposing assumptions on empirical parameters such as the cone factor and strain rate correction factor, making the interpretation of PFFP data more user-friendly, especially for users without a geotechnical engineering background.

A key question about this study is whether the model can be applied to other sites, especially those with different sediment characteristics than the training data, because data from only three main locations were used. An attempt was made to answer this question by testing the model in a site that is situated in a different geographical region. Nevertheless, if the input deceleration curve is very different from what the model was trained on, the model will have a higher degree of uncertainty in its predictions. This uncertainty, known as epistemic uncertainty (Der Kiureghian and Ditlevsen 2009), occurs because the model lacks sufficient data to make accurate predictions for these new conditions. By adding more data to the training set, this uncertainty can be reduced, which eventually will make the model more accurate. The model is

Class 2, and vice versa. There was a similar pattern for Classes 3 and 4. This observation indicates that the PFFP exhibits distinctly different deceleration curves for cohesionless (Classes 1 and 2) versus cohesive (Classes 3 and 4) sediments, which is consistent with the findings of previous studies (Mulukutla et al. 2011; Jaber and Stark 2023). However, the distinction is less clear between Class 1 and Class 2, and between Class 3 and Class 4. The primary differentiator between Classes 1 and 2 is the fines content in the sediment, with a threshold set at 12%. For example, sediments containing 11% and 13% fines may exhibit deceleration curves that are not significantly different, yet they are classified into different categories. Additionally, some of the misclassification also may result from the fact that these classes are based on the USCS classification system, which has its own limitations. The Revised Soil Classification System (RSCS) proposed by Park and Santamarina (2017) has notable advantages over the traditional Unified Soil designed to easily incorporate additional data, which means that it can be improved and made more reliable for a wider range of conditions in the future.

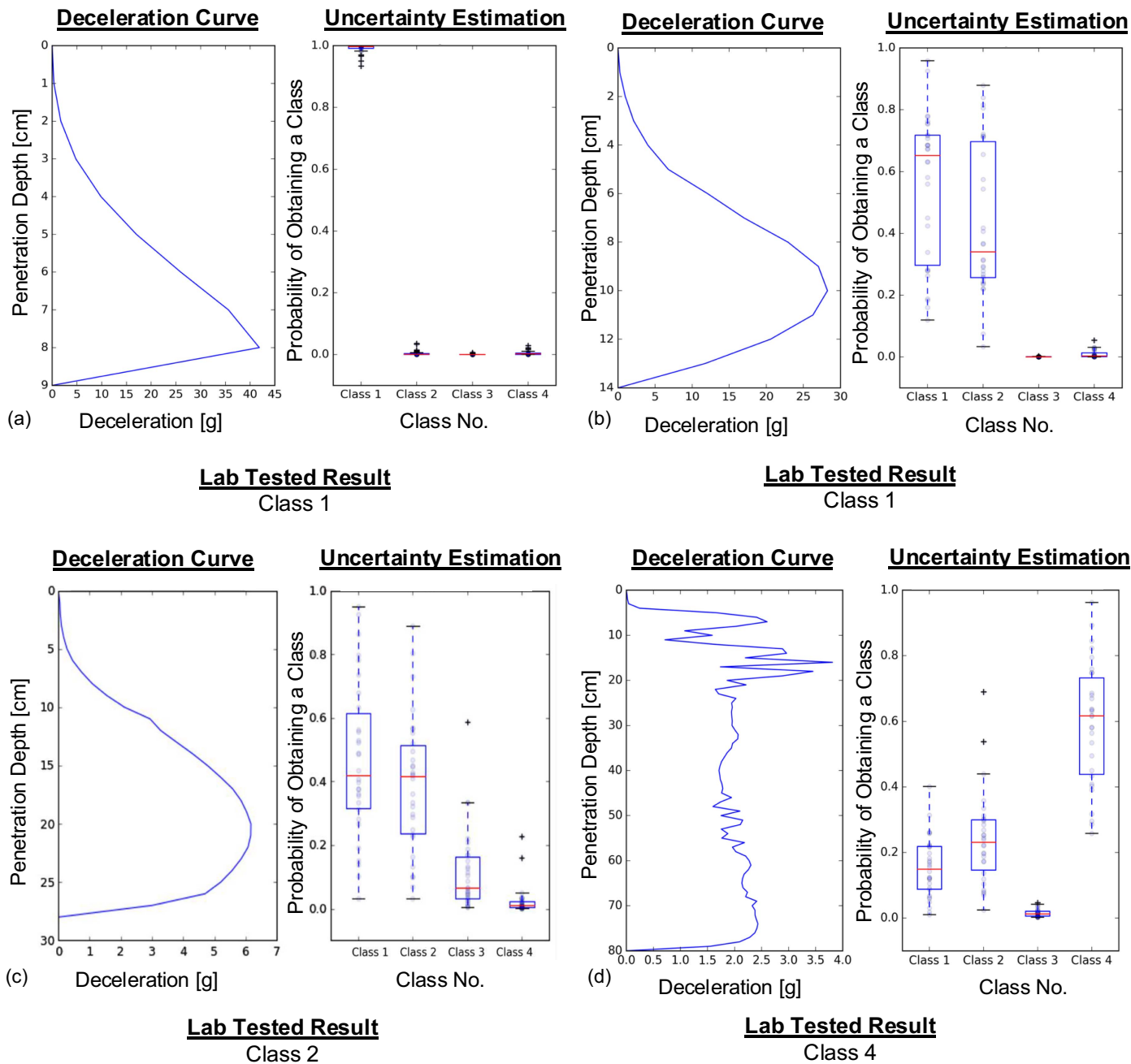

**Fig. 11.** Different types of prediction cases obtained from the proposed model: (a) prediction case showing little to no uncertainty in prediction; (b) prediction case showing uncertainty between two cohesionless classes; (c) prediction case showing uncertainty among three classes; and (d) prediction case for a deceleration curve likely affected by shell hash and clams.

However, there is another type of uncertainty, which is called aleatoric or inherent variability. The inferences made from the PFFP, as proposed in this paper, assume a uniform sediment. Deviations from this assumption, which can result from inherent variability in the physical characteristics of the sediment, would manifest as aleatory variability. For example, information about the mineralogy of the sediment, the presence of layering in the sediments, knowledge of biological elements such as burrowing infauna or biofilms can affect the associated uncertainty of the prediction. Future studies could focus on integrating these additional physical and biological parameters into the model to better account for aleatoric uncertainty and enhance prediction accuracy.

The predicted classes have strong potential to be interpreted further to provide approximate sediment behaviors regarding bed erodibility. Erodibility estimates involve calculating critical bed shear stress ($\tau_{cr}$), which is based on sediment properties, indicating sediment strength against hydrodynamic forces (Soulsby 1997). For cohesionless sediments, $\tau_{cr}$ often is estimated using median grain size ($d_{50}$). With cohesive components, $\tau_{cr}$ increases until a mineralogy-dependent threshold, after which it decreases as sediments become fully cohesive (Whitehouse 2000). Applying this to our classification, Class 1 (cohesionless, low plasticity) should have higher erodibility than Class 2 (cohesionless, some plasticity).

Shafii et al. (2023) presented a chart for obtaining the range of erodibility based on the USCS classification. According to that chart, Class 1 of our classification should fall in the very high to high erodibility zone, whereas Class 2 will fall in the high to medium erodibility zone. Classes 3 (low-plasticity cohesive) and 4 (high-plasticity cohesive) exhibit cohesive behavior, with significant factors such as plasticity, organic content, and biological activity influencing erodibility. Kamphuis and Hall (1983) suggested that with increasing plasticity, the critical bed stress also increases. From this point of view, Class 4 should have lower erodibility than Class 3 if they both have similar fines content because Class 4 has higher plasticity. Both Class 3 and 4 fall within high to low erodibility according to Shafii et al. (2023), with variability due to additional factors mentioned previously. Sediment transport due to erosion follows three modes: noncohesive (bedload), transitional (mix of bedload and suspended load), and cohesive (suspended load). In our classification, Class 1 should align with noncohesive to transitional modes, Class 2 with transitional to cohesive modes, and Classes 3 and 4 with cohesive sediment transport.

Another sediment behavior into which the predicted classes can provide insight susceptibility to wave-induced liquefaction, which significantly depends on sediment composition. Under large, steep waves in relatively shallow water, high-permeability clean sands containing small amounts of gas can be prone to momentary wave-induced liquefaction because the gas causes the pore water–gas mixture to become compressible (Michallet et al. 2009). Under less-extreme wave conditions, residual liquefaction can occur in semipermeable sand–mud mixtures due to the build-up of excess pore pressure over a series of waves. Experiments with kaolinite–fine sand mixtures and illite–fine sand mixtures demonstrated that low but nonzero clay content (≥ 0.5%) increases susceptibility to residual liquefaction in semipermeable, loosely compacted sands, whereas higher clay content (typically over 10%–30%, depending on clay minerals) can decrease susceptibility as permeability and compressibility decrease (Zhang et al. 2020). This suggests that Class 1 may be the only class that is susceptible to momentary wave-induced liquefaction, and Classes 1 and 2 sediments may be prone to wave-induced residual liquefaction to differing degrees, depending on their compaction and fines content. Predominantly fine-grained classes (Class 3 and 4) may be minimally susceptible to wave-induced liquefaction because their permeability is too low.

Object burial is an additional important process that researchers work to predict using sediment behavior. It is a complex process influenced by bearing failure, scouring, consolidation, and fluidization. Scour burial within fully cohesionless sediments (Class 1) has been well-documented (Traykovski et al. 2007; Trembanis et al. 2007; Whitehouse 1998). Friedrichs et al. (2016) found that, acting alone, scour burial typically is limited to lowering the base of the object to 1–1.5 diameters below the far-field seabed surface in cohesionless sediments (Class 1 and Class 2). At that point, the action of scour simply keeps the object from being fully covered. Conversely, cohesive sediments (Class 3 and 4) tend to exhibit notably higher burial depths due to impact burial followed by consolidation (Trembanis and DuVal 2021). Overall, due to the combined effects of scour, impact, and consolidation burial, objects in Class 1 and Class 2 sediments should often be partially exposed, whereas Class 3 and Class 4 sediments more typically should be fully buried.

The uncertainty estimates from the proposed model can be interpreted further to reveal insights into sediment behaviors. For example, the sediment depicted in Fig. 11(b) is likely Class 1 or Class 2. As discussed previously, this suggests that the sediment will have very high to medium erodibility, as approximated by Shafii et al. (2023). Factors such as plasticity, organic content, and biological activity will be less significant to sediment properties because these become relatively more important in fine-grained cohesive sediments (Class 3 and 4).

## Conclusion

This study introduces a novel predictive model for classifying sediment behavior using data from portable free-fall penetrometer deployments. By integrating random forest and 1D Bayesian convolutional neural network algorithms, the model is capable of producing satisfactory classification of surficial sediments. The four sediment classes identified offer valuable insights into sediment characteristics essential for coastal engineering applications. The model not only categorizes sediments but also provides uncertainty estimates, offering a probabilistic range for each class prediction. This enhances decision-making processes by allowing users to assess potential risks and interpret sediment behavior more comprehensively. Importantly, the model avoids assumptions about parameters such as the cone factor and strain rate correction factor, making it accessible to those without specialized geotechnical expertise.

However, the study has limitations, particularly the reliance on PFFP measurements from only three locations for training and testing the model, which raises concerns about its generalizability. Nevertheless, the model's design allows for the incorporation of additional data, suggesting that expanding the training data set with more diverse sites and environmental conditions could significantly enhance its applicability. Future research should focus on expanding the data set and refining the model to improve its robustness and accuracy further. Exploring quantitative approaches for predicting erodibility or object burial using the full potential of PFFP data sets also could be beneficial. Such advancements would greatly enhance sediment mapping processes and engineering decision-making in nearshore environments, ultimately contributing to more effective coastal management and infrastructure development.

## Data Availability Statement

Some or all data, models, or code generated or used during the study are available in a repository online in accordance with funder data retention policies: https://github.com/Rejwan05/PFFP_Prediction.git.

## Acknowledgments

Funding for this work was provided by the Strategic Environmental Research and Development Program (SERDP) through Grant no. MR21-1265. The authors thank Eric Hunstein, Saurav Shrestha, Elise Hummel, Nick Brilli, and Jonathan Moore (all from Virginia Tech), and Chesna Cox (DISL) for their assistance with field measurements. The authors also thank Dr. Bruce Hatcher (Cape Breton University) for assistance and guidance in conducting field measurements in Sydney, Nova Scotia. While preparing this manuscript, ChatGPT was used for paraphrasing the writing.

## Author Contributions

Md Rejwanur Rahman: Data curation; Formal analysis; Investigation; Methodology; Validation; Visualization; Writing – original draft; Writing – review and editing. Adrian Rodriguez-Marek: Conceptualization; Funding acquisition; Supervision; Writing – review and editing. Nina Stark: Conceptualization; Funding acquisition;

Supervision; Writing – review and editing. Grace Massey: Funding acquisition; Supervision; Writing – review and editing. Carl Friedrichs: Funding acquisition; Supervision; Writing – review and editing. Kelly M. Dorgan: Funding acquisition; Supervision; Writing – review and editing.

## References

Akal, T., and R. Stoll. 1995. "An expendable penetrometer for rapid assessment of seafloor parameters." In Vol. 3 of *Proc., OCEANS'95 MTS/IEEE, 'Challenges of Our Changing Global Environment', Conf.*, 1822–1826. New York: IEEE. https://doi.org/10.1109/OCEANS.1995.528858.

Albatal, A., and N. Stark. 2017. "Rapid sediment mapping and in situ geotechnical characterization in challenging aquatic areas." *Limnol. Oceanogr. Methods* 15 (8): 690–705. https://doi.org/10.1002/lom3.10192.

Albatal, A., N. Stark, and B. Castellanos. 2020. "Estimating in situ relative density and friction angle of nearshore sand from portable free-fall penetrometer tests." *Can. Geotech. J.* 57 (1): 17–31. https://doi.org/10.1139/cgj-2018-0267.

ASTM. 2017. *Standard practice for classification of soils for engineering purposes (unified soil classification system)*. ASTM D2487-17. West Conshohocken, PA: ASTM.

Aubeny, C., and H. Shi. 2006. "Interpretation of impact penetration measurements in soft clays." *J. Geotech. Geoenviron. Eng.* 133 (7): 767–781. https://doi.org/10.1061/(ASCE)1090-0241(2006)132:6(770).

Bilici, C., N. Stark, C. Friedrichs, and G. Massey. 2019. "Coupled sedimentological and geotechnical data analysis of surficial sediment layer characteristics in a tidal estuary." *Geo-Mar. Lett.* 39 (3): 175–189. https://doi.org/10.1007/s00367-019-00565-3.

Blundell, C., J. Cornebise, K. Kavukcuoglu, and D. Wierstra. 2015. "Weight uncertainty in neural network." Preprint, submitted May 20, 2015. https://arxiv.org/abs/1505.05424.

Breiman, L. 2001. "Random forests." *Mach. Learn.* 45 (1): 5–32. https://doi.org/10.1023/A:1010933404324.

Bridle, J. 1989. "Training stochastic model recognition algorithms as networks can lead to maximum mutual information estimation of parameters." In Vol. 2 of *Proc., Advances in Neural Information Processing Systems*. San Francisco: Morgan-Kaufmann.

Cetin, K., and C. Ozan. 2009. "CPT-based probabilistic soil characterization and classification." *J. Geotech. Geoenviron. Eng.* 135 (1): 84–107. https://doi.org/10.1061/(ASCE)1090-0241(2009)135:1(84).

Chow, S., C. O'Loughlin, and M. Randolph. 2014. "Soil strength estimation and pore pressure dissipation for free-fall piezocone in soft clay." *Géotechnique* 64 (10): 817–827. https://doi.org/10.1680/geot.14.P.107.

Chow, S., C. O'Loughlin, D. White, and M. Randolph. 2017. "An extended interpretation of the free-fall piezocone test in clay." *Géotechnique* 67 (12): 1090–1103. https://doi.org/10.1680/jgeot.16.P.220.

Der Kiureghian, A., and O. Ditlevsen. 2009. "Aleatory or epistemic? Does it matter?" *Struct. Saf.* 31 (2): 105–112. https://doi.org/10.1016/j.strusafe.2008.06.020.

Douglas, B., and R. Olsen. 1981. "Soil classification using electric cone penetrometer." In *Proc., Symp. on Cone Penetration Testing and Experience, Geo-technical Engineering Division*. Reston, VA: ASCE.

Friedrichs, C., S. Rennie, and A. Brandt. 2016. "Self-burial of objects on sandy beds by scour: A synthesis of observations." In *Proc., 8th Int. Conf. on Scour and Erosion*, 179. Boca Raton, FL: CRC Press.

Gal, Y., and Z. Ghahramani. 2016. "Bayesian convolutional neural networks with Bernoulli approximate variational inference." Preprint, submitted June 6, 2015. https://arxiv.org/abs/1506.02158.

Goodfellow, I., Y. Bengio, A. Courville, and Y. Bengio. 2016. *Deep learning*. Cambridge, MA: MIT Press.

Graves, A. 2011. "Practical variational inference for neural networks." In Vol. 24 of *Proc., Advances in Neural Information Processing Systems*. Red Hook, NY: Curran Associates.

He, H., Y. Bai, E. Garcia, and S. Li. 2008. "ADASYN: Adaptive synthetic sampling approach for imbalanced learning." In *Proc., IEEE Int. Joint Conf. on Neural Networks (IEEE World Congress on Computational Intelligence)*, 1322–1328. New York: IEEE. https://doi.org/10.1109/IJCNN.2008.4633969.

Hunstein, E., N. Stark, and A. Rodriguez-Marek. 2023. "Finding the mudline: Automatization of seabed impact selection for a portable free-fall penetrometer." *J. Geotech. Geoenviron. Eng.* 149 (12): 06023008. https://doi.org/10.1061/JGGEFK.GTENG-11550.

Jaber, R., and N. Stark. 2023. "Geotechnical properties from portable free fall penetrometer measurements in coastal environments." *J. Geotech. Geoenviron. Eng.* 149 (12): 04023120. https://doi.org/10.1061/JGGEFK.GTENG-11013.

Jung, B., P. Gardoni, and A. Biscontin. 2008. "Probabilistic soil identification based on cone penetration tests." *Géotechnique* 58 (7): 591–603. https://doi.org/10.1680/geot.2008.58.7.591.

Kamphuis, J., and K. Hall. 1983. "Cohesive material erosion by unidirectional current." *J. Hydraul. Eng.* 109 (1): 49–61. https://doi.org/10.1061/(ASCE)0733-9429(1983)109:1(49).

Kiranyaz, S., O. Avci, O. Abdeljaber, T. Ince, M. Gabbouj, and D. Inman. 2021. "1D convolutional neural networks and applications: A survey." *Mech. Syst. Signal Process.* 151 (Apr): 107398. https://doi.org/10.1016/j.ymssp.2020.107398.

Krogh, A. 2008. "What are artificial neural networks?" *Nat. Biotechnol.* 26 (2): 195–197. https://doi.org/10.1038/nbt1386.

Michallet, H., M. Mory, and I. Piedra-Cueva. 2009. "Wave-induced pore pressure measurements near a coastal structure." *J. Geophys. Res. Oceans* 114 (C6). https://doi.org/10.1029/2008JC005071.

Mitchell, J. K., and K. Soga. 2005. *Fundamentals of soil behavior*. 3rd ed. Hoboken, NJ: John Wiley & Sons.

Müller, A. C., and S. Guido. 2016. *Introduction to machine learning with Python: A guide for data scientists*. Sebastopol, CA: O'Reilly Media.

Mulukutla, G., L. Huff, J. Melton, K. Baldwin, and L. Mayer. 2011. "Sediment identification using free fall penetrometer acceleration-time histories." *Mar. Geophys. Res.* 32 (3): 397–411. https://doi.org/10.1007/s11001-011-9116-2.

Mumtaz, B., N. Stark, and S. Brizzolara. 2018. "Pore pressure measurements using a portable free fall penetrometer." In *Proc., Cone Penetration Testing 2018 (CPT'18)*, London: CRC Press.

Murphy, K. P. 2022. *Probabilistic machine learning: An introduction*. Cambridge, MA: MIT Press.

Nair, V., and G. Hinton. 2010. "Rectified linear units improve restricted Boltzmann machines." In *Proc., 27th Int. Conf. on Machine Learning*, 807–814. Toronto, ON, Canada: Univ. of Toronto.

Olsen, R., and J. Mitchell. 1995. "CPT stress normalization and prediction of soil classification." In Vol. 2 of *Proc., Int. Symp. on Cone Penetration Testing, CPT'95*, 257–262. Linköping, Sweden: Swedish Geotechnical Society.

Park, J., and J. C. Santamarina. 2017. "Revised soil classification system for coarse-fine mixtures." *J. Geotech. Geoenviron. Eng.* 143 (8): 04017039. https://doi.org/10.1061/(ASCE)GT.1943-5606.0001705.

Pedregosa, F., et al. 2011. "Scikit-learn: Machine learning in python." *J. Mach. Learn. Res.* 12 (Nov): 2825–2830.

Pradhan, T. B. S. 1998. "Soil identification using piezocone data by fuzzy method." *Soils Found.* 38 (1): 255–262. https://doi.org/10.3208/sandf.38.255.

Rahman, M. R. 2025. "Probabilistic classification of near-surface shallow-water sediments using a portable free-fall penetrometer." *Zenodo*. https://doi.org/10.5281/zenodo.14739151.

Robertson, P. 1990. "Soil classification using the cone penetration test." *Can. Geotech. J.* 27 (1): 151–158. https://doi.org/10.1139/t90-014.

Shafii, I., Z. Medina-Cetina, A. Shidlovskaya, and J.-L. Briaud. 2023. "Relationship between soil erodibility and soil properties." *J. Geotech. Geoenviron. Eng.* 149 (1): 04022121. https://doi.org/10.1061/(ASCE)GT.1943-5606.0002915.

Soulsby, R. 1997. *Dynamics of marine sands*. London: Thomas Telford.

Stark, N., G. Coco, K. R. Bryan, and A. Kopf. 2012. "In-situ geotechnical characterization of mixed-grain-size bedforms using a dynamic penetrometer." *J. Sediment. Res.* 82 (7): 540–544. https://doi.org/10.2110/jsr.2012.45.

Stark, N., K. M. Dorgan, N. C. Brilli, M. R. Frey, C. Cox, and J. Calantoni. 2024. "Initial observations of the impacts of infauna on portable free fall penetrometer measurements in sandy parts of mobile bay." *Acta Geotech.* 19 (3): 1251–1265. https://doi.org/10.1007/s11440-024-02241-y.

Traykovski, P., M. Richardson, L. Mayer, and J. Irish. 2007. "Mine burial experiments at the Martha's vineyard coastal observatory." *IEEE J. Ocean. Eng.* 32 (1): 150–166. https://doi.org/10.1109/JOE.2007.890956.

Trembanis, A., and C. DuVal. 2021. *Further examining the role of cohesive sediments in munitions mobility through additional infield deployment of smart munitions and application of a SERDP-developed penetrometer.* Final Rep. No. MR20-1480. Alexandria, VA: Strategic Environmental Research and Development Program.

Trembanis, A., C. Friedrichs, M. Richardson, P. Traykovski, P. Howd, P. Elmore, and T. Wever. 2007. "Predicting seabed burial of cylinders by wave-induced scour: Application to the sandy inner shelf off Florida and Massachusetts." *IEEE J. Ocean. Eng.* 32 (1): 167–183. https://doi.org/10.1109/JOE.2007.890958.

Whitehouse, R. 1998. *Scour at marine structures: A manual for practical applications.* Leeds, UK: Emerald Publishing.

Whitehouse, R. 2000. *Dynamics of estuarine muds: A manual for practical applications.* London: Thomas Telford. https://doi.org/10.1680/doem.28647.

Zhang, J., Q. Jiang, D. Jeng, C. Zhang, X. Chen, and L. Wang. 2020. "Experimental study on mechanism of wave-induced liquefaction of sand-clay seabed." *J. Mar. Sci. Eng.* 8 (2): 66. https://doi.org/10.3390/jmse8020066.

Zhang, Z., and M. T. Tumay. 1999. "Statistical to fuzzy approach toward CPT soil classification." *J. Geotech. Geoenviron. Eng.* 125 (3): 179–186. https://doi.org/10.1061/(ASCE)1090-0241(1999)125:3(179).